\documentclass[letterpaper, 10 pt, conference]{ieeeconf}  

\IEEEoverridecommandlockouts                              

\usepackage{graphicx}
\usepackage{amsmath}
\usepackage{amssymb}
\let\labelindent\relax
\usepackage{enumitem}
\usepackage[bookmarks=true]{hyperref}
\hypersetup{
  colorlinks   = true, 
  urlcolor     = blue, 
  linkcolor    = blue, 
  citecolor   = red 
}
\usepackage{graphicx}
\usepackage{listings}
\usepackage{comment}
\usepackage[ruled,linesnumbered,vlined]{algorithm2e}

\usepackage{amsthm}
\usepackage[short,c2,nocomma]{optidef}
\usepackage{svg}
\usepackage{adjustbox}

\usepackage[title]{appendix}
\usepackage{wrapfig}
\usepackage{cite}
\usepackage{float}
\usepackage{mathtools}
\usepackage{placeins}

{\itshape}{\rmfamily}
{\itshape}{\rmfamily}
{\bfseries}{\itshape}
\newtheorem{definition}{Definition}{\bfseries}{\itshape}
{\itshape}{\rmfamily}
{\itshape}{\rmfamily}
{\bfseries}{\itshape}
{\itshape}{\rmfamily}
{\itshape}{\rmfamily}
{\itshape}{\rmfamily}
{\bfseries}{\itshape}
{\itshape}{\rmfamily}
{\itshape}{\rmfamily}
{\itshape}{\rmfamily}
\newtheorem{theorem}{Theorem}{\bfseries}{\itshape}

\def\Qfree{\mathcal{Q}_{free}}

\makeatletter
\newcommand{\subalign}[1]{%
  \vcenter{%
    \Let@ \restore@math@cr \default@tag
    \baselineskip\fontdimen10 \scriptfont\tw@
    \advance\baselineskip\fontdimen12 \scriptfont\tw@
    \lineskip\thr@@\fontdimen8 \scriptfont\thr@@
    \lineskiplimit\lineskip
    \ialign{\hfil$\m@th\scriptstyle##$&$\m@th\scriptstyle{}##$\hfil\crcr
      #1\crcr
    }%
  }%
}
\makeatother

\definecolor{DarkGreen}{RGB}{1,150,32}

\SetKwFunction{UnvisitedTargets}{UnvisitedTargets}
\SetKwFunction{UnvisitedWindows}{UnvisitedWindows}
\SetKwFunction{VisitedTargets}{VisitedTargets}
\SetKwFunction{IsInf}{IsInf}
\SetKwFunction{NextWindowIndex}{NextWindowIndex}
\SetKwFunction{EquivalentPartialTours}{EquivalentPartialTours}
\SetKwFunction{Traj}{Traj}
\SetKwFunction{Tar}{Tar}
\SetKwFunction{Key}{Key}
\SetKwFunction{LTraj}{LTraj}
\SetKwFunction{LTime}{LTime}

\title{\LARGE \bf
A Complete and Bounded-Suboptimal Algorithm for a Moving Target Traveling Salesman Problem with Obstacles in 3D*}

\author{Anoop Bhat$^{1}$ and Geordan Gutow$^{1}$ and Bhaskar Vundurthy$^{1}$ and\\Zhongqiang Ren$^{2}$ and Sivakumar Rathinam$^{3}$ and Howie Choset$^{1}$
\thanks{$^{1}$Robotics Institute at Carnegie Mellon University, 5000 Forbes Ave., Pittsburgh, PA 15213, USA. Emails: \{agbhat,
ggutow, pvundurt, choset\}@andrew.cmu.edu. This research was supported by an appointment to the Intelligence Community Postdoctoral Research Fellowship Program.}%
\thanks{$^{2}$UM-SJTU Joint Institute and Department of Automation at Shanghai Jiao Tong University, Shanghai, China. Email: zhongqiang.ren@sjtu.edu.cn}%
\thanks{$^{3}$Department of Mechanical Engineering and Department of Computer Science and Engineering at Texas A\&M University, College Station, TX 77843. Email: srathinam@tamu.edu}%
}

\begin{document}

\maketitle
\thispagestyle{plain}
\pagestyle{plain}

\begin{abstract}
The moving target traveling salesman problem with obstacles (MT-TSP-O) seeks an obstacle-free trajectory for an agent that intercepts a given set of moving targets, each within specified time windows, and returns to the agent's starting position. Each target moves with a constant velocity within its time windows, and the agent has a speed limit no smaller than any target's speed. We present FMC*-TSP, the first complete and bounded-suboptimal algorithm for the MT-TSP-O, and results for an agent whose configuration space is $\mathbb{R}^3$. Our algorithm interleaves a high-level search and a low-level search, where the high-level search solves a generalized traveling salesman problem with time windows (GTSP-TW) to find a sequence of targets and corresponding time windows for the agent to visit. Given such a sequence, the low-level search then finds an associated agent trajectory. To solve the low-level planning problem, we develop a new algorithm called FMC*, which finds a shortest path on a graph of convex sets (GCS) via implicit graph search and pruning techniques specialized for problems with moving targets. We test FMC*-TSP on 280 problem instances with up to 40 targets and demonstrate its smaller median runtime than a baseline based on prior work.
\end{abstract}

\section{INTRODUCTION}\label{sec:intro}
\begin{figure}
    \centering
    \includegraphics[width=0.47\textwidth]{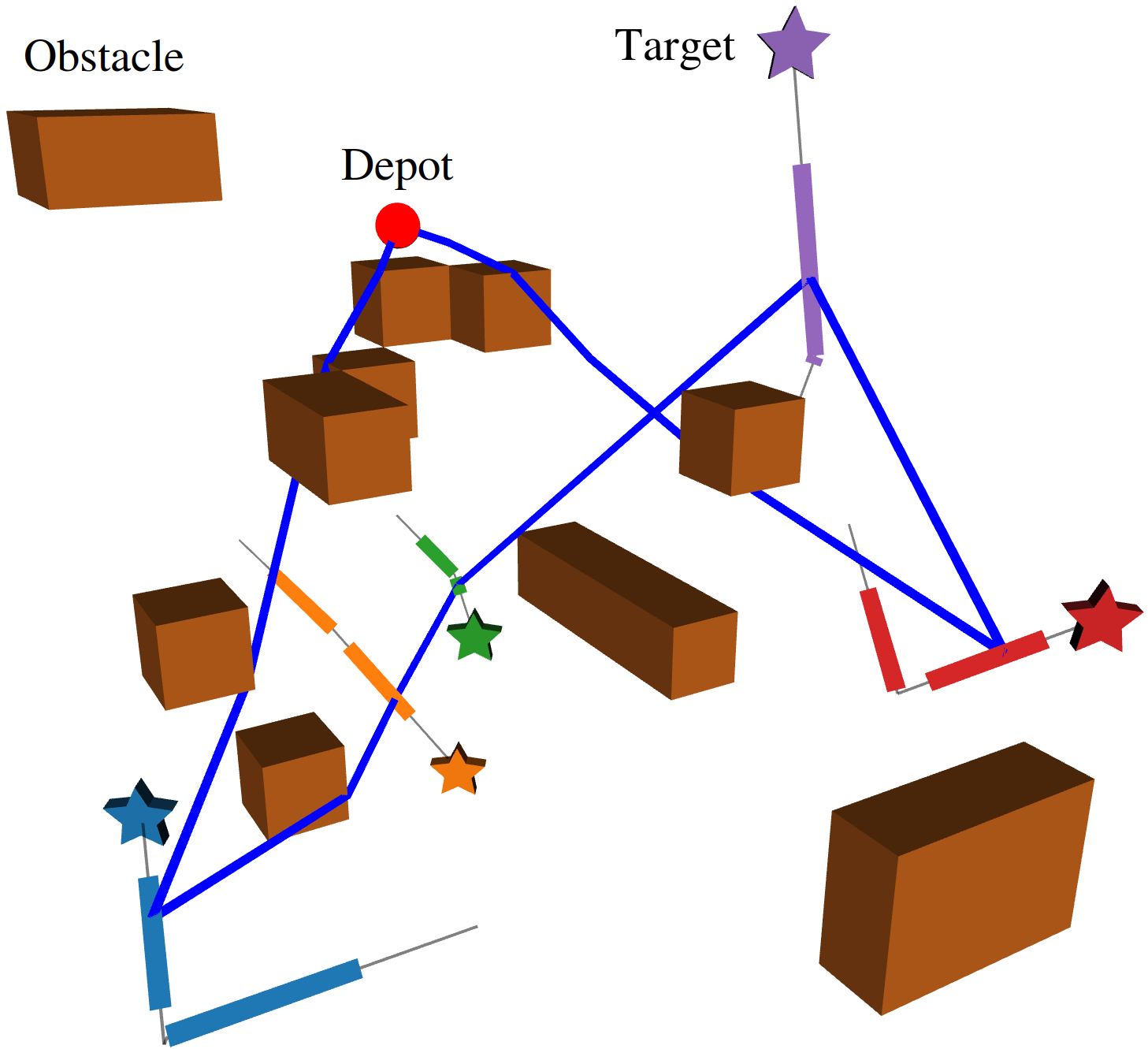}
    \vspace{-0.3cm}
    \caption{Targets depicted as stars move along piecewise-linear trajectories. Portions of a target's trajectory within time windows are highlighted in bold colors. Agent trajectory shown in blue begins and ends at depot, intercepting each target within one of its time windows while avoiding obstacles.}
    \label{fig:intro_fig}
    \vspace{-0.6cm}
\end{figure}

Visiting moving targets in environments with obstacles is necessary in applications ranging from underway replenishment of naval ships~\cite{brown2017scheduling} to delivering spare parts to spacecraft on-orbit~\cite{bourjolly2006orbit}. The shortest path problem of visiting multiple targets by an agent is often modeled as a traveling salesman problem (TSP) in the literature \cite{cook2011traveling, gutin2006traveling}. Given the cost of travel between any pair of targets, the TSP aims to find a sequence of targets to visit such that each target is visited exactly once and the sum of the travel costs for the agent is minimized. The moving target TSP (MT-TSP) is a generalization of the TSP where the targets are mobile, and may only be visited within specific time windows. The MT-TSP seeks a sequence of targets as well as a trajectory through space for an agent that intercepts each target. When the agent must avoid static obstacles, we have the moving target TSP with obstacles (MT-TSP-O) (refer to Fig. \ref{fig:intro_fig}).

As with other motion planning problems, two desirable properties in an algorithm for the MT-TSP-O are completeness and optimality. No existing MT-TSP-O algorithm has both properties. Even finding a feasible solution to the MT-TSP-O is NP-complete~\cite{savelsbergh1985local} due to the presence of time windows, making completeness difficult to guarantee. Furthermore, since the MT-TSP-O generalizes the TSP, solving the MT-TSP-O optimally is NP-hard. Our prior work~\cite{bhat2024AComplete} developed a complete algorithm for the MT-TSP-O in $\mathbb{R}^2$ using a generalization of a visibility graph. The completeness of~\cite{bhat2024AComplete} is limited to the plane, where a shortest path in a visibility graph is also a shortest path in continuous space. This property fails in three dimensions\footnote{As in figure 1, the agent's path from the blue target to the green target wraps around an edge of an obstacle rather than a vertex. A visibility graph only considers paths that move between obstacle vertices.}. Note that with MT-TSP-O, path length affects completeness because of timing constraints: if an agent takes an excessively long path from its starting point to a destination point on a target's trajectory, the agent may not reach its destination point by the time the target gets there, even if doing so is possible.

In this paper, we propose the first complete and bounded-suboptimal algorithm for the MT-TSP-O in three-dimensions. Our algorithm interleaves a high-level search for a sequence of targets and associated time windows, and a low-level search for a trajectory intercepting a given target/time window sequence. The low-level search uses a graph of convex sets (GCS)~\cite{marcucci2024shortest}, rather than the visibility-based graph in~\cite{bhat2024AComplete}, because the visibility graph approach is limited to the plane. In particular, the low-level search extends existing techniques for implicit graph search on a GCS~\cite{natarajan2024ixg, chia2024gcsforwardheuristicsearch}, using focal search~\cite{pearl1982studies} to enforce bounded-suboptimality and additional limits on arrival times to intermediate targets to prune suboptimal solutions. Therefore, one of our contributions is a new GCS-based algorithm for intercepting a given sequence of moving targets in minimum-time, which we call FMC* (\underline{F}ocal search for \underline{M}oving targets on a graph of \underline{C}onvex sets). We call our overall algorithm FMC*-TSP. We test FMC*-TSP against a baseline based on prior work that samples the trajectories of targets into points. The baseline is not complete or bounded-suboptimal, and often needs a large number of samples to find a solution. As we increase the number of targets, FMC*-TSP often finds solutions for the MT-TSP-O more quickly compared to the baseline while satisfying a specified suboptimality bound.

\section{RELATED WORK}
Several algorithms exist for the MT-TSP without obstacles, ranging from complete and optimal methods \cite{philip2024mixedinteger, stieber2022DealingWithTime,stieber2022multiple,philip2025mixed} to incomplete and suboptimal heuristic methods~\cite{bourjolly2006orbit, choubey2013moving, groba2015solving, ucar2019meta, jiang2005tracking, de2019experimental, 6760409, marlow2007travelling, wang2023moving}. The complete and optimal methods assume trajectories of targets are linear or piecewise-linear; our work makes the same piecewise-linear assumption. In the presence of obstacles, there are two related works. \cite{Li2019RendezvousPlanning} represents trajectories of targets using sample points and requires a straight line agent trajectory between each sample point: this method is neither complete nor optimal. Our prior work~\cite{bhat2024AComplete} develops a complete (but not an optimal) algorithm for the MT-TSP-O, for planar instances with piecewise-linear target trajectories.

A complete and optimal method of planning trajectories in non-planar environments is to plan a shortest path on a graph of convex sets (GCS)~\cite{marcucci2024shortest, marcucci2023motion}. \cite{marcucci2024shortest} finds the shortest path on a GCS via a mixed integer convex program (MICP), which can be solved exactly to guarantee completeness and optimality, or solved via the relaxation-then-rounding method from~\cite{marcucci2023motion} that sacrifices guarantees to improve runtime. IxG*~\cite{natarajan2024ixg} and GCS*~\cite{chia2024gcsforwardheuristicsearch} replace the MICP in~\cite{marcucci2024shortest} with implicit graph search, allowing operations on larger graphs. The low-level planner in our work also implicitly searches a GCS, but exploits the structure of minimum-time problems with moving targets and time windows to accelerate the search.

\section{PROBLEM SETUP}\label{sec:problem_setup}
Let $\Qfree$ be the obstacle-free configuration space in $\mathcal{Q} = \mathbb{R}^3$. For some final time $t_f$, let the agent's trajectory be $\tau_a: [0, t_f] \rightarrow \mathcal{Q}$. For all $t \in [0, t_f]$, we require $\tau_a(t) \in \Qfree$, $\tau_a(0)=p_d$ (the position of the depot), and that $\tau_a$ is \textit{speed admissible}, $i.e.$ it never exceeds the agent's speed limit $v_{max}$.

Denote the set of moving targets as $\mathcal{I} = \{1, 2, \dots, |\mathcal{I}|\}$. Each target $i \in \mathcal{I}$ is associated with a set of time windows, $\mathcal{W}_i = \{[\underline{t}_{i}^1, \bar{t}_{i}^1], [\underline{t}_{i}^2, \bar{t}_{i}^2], \dots, [\underline{t}_{i}^{|\mathcal{W}_i|}, \bar{t}_{i}^{|\mathcal{W}_i|}]\}$. $\underline{t}_{i}^j$ and $\bar{t}_{i}^j$ denote the start and end time of target $i$'s $j$th time window respectively. Let the trajectory of target $i$ be $\tau_i: \mathbb{R} \rightarrow \mathcal{Q}$. We assume within each of target $i$'s time windows, $\tau_i$ lies in $\Qfree$ and has speed no larger than $v_{max}$. We also assume $\tau_i$ has constant velocity within each time window in $\mathcal{W}_i$, though the velocity may differ from one window to another.

We define a \textit{target-window} $u_{i}^j$ as the pairing of target $i$ with its $j$th time window, i.e. $u_{i}^j = (i, [\underline{t}_i^j, \overline{t}_i^j])$. We also define a target-window associated with the depot, $u_d = u_0^1 = (0, [0, \infty))$. Here, we treat the depot as a fictitious target 0 with $\tau_0(t) = p_d$ for all $t$, and with $\mathcal{W}_0 = \{[0, \infty)\}$. Let $\mathcal{T}_i = \{i\} \times \mathcal{W}_i$ be the set of target-windows for target $i$. For each target-window $u_i^j$, we let $\Tar(u_i^j) = i$, $\underline{t}(u_i^j) = \underline{t}_{i}^{j}$, $\bar{t}(u_i^j) = \bar{t}_{i}^{j}$, and $\Traj(u_i^j) = \tau_i$. An agent trajectory $\tau_a$ \textit{intercepts} target-window $u$ at time $t$ if $u_a(t) = \Traj(u)(t)$.

A \textit{tour} is a sequence of target-windows containing exactly one target-window per target, beginning and ending with $u_d$. A \textit{partial tour} is a sequence of target-windows beginning with $u_d$ and containing at most one target-window per target, not necessarily ending with $u_d$. For a partial tour $\Gamma$, $|\Gamma|$ is the length, $\Gamma[i]$ is the $i$th element\footnote{We use 1-based indexing, i.e. the first element of $\Gamma$ is $\Gamma[1]$}, and $\Gamma[-1]$ is the final element. Additionally, the length-$i$ \textit{prefix} of $\Gamma$ is $\Gamma[:i] = (\Gamma[1], \Gamma[2], \dots, \Gamma[i])$. A partial tour $\Gamma$ \textit{visits} target $i$ if for some $u \in \mathcal{T}_i$, $u \in \Gamma$. An agent trajectory $\tau_a$ \textit{executes} a partial tour $\Gamma$ if $\tau_a$ intercepts the target-windows in $\Gamma$ in order. The MT-TSP-O seeks a tour $\Gamma$, an agent trajectory $\tau_a$, and a final time $t_f$ such that $\tau_a$ executes $\Gamma$, and $t_f$ is minimized. In this work, we assume that we are given a suboptimality factor $w$, such that if the optimal value of $t_f$ is $t_f^*$, then our returned solution ($\Gamma, \tau_a, t_f)$ satisfies $t_f \leq wt_f^*$.

\section{FMC*-TSP ALGORITHM}
An overview of FMC*-TSP is shown in Alg. \ref{alg:fmc_star_tsp}. FMC*-TSP searches for a tour on a target-window graph (Section \ref{subsec:twg}), where the nodes are target-windows, and an edge connects target-window $u$ to $v$ if the agent can intercept $u$, then $v$, in the absence of obstacles. The search for a tour is posed as a generalized traveling salesman problem with time windows (GTSP-TW) (Section \ref{subsec:gtsptw}). For every tour generated by the GTSP-TW solver, FMC*-TSP performs a low-level search on a GCS for a trajectory executing the tour (Sections \ref{subsec:low_level_search} to \ref{subsec:fmc_star}). The low-level search  updates an incumbent solution $(\Gamma^{inc}, \tau_a^{inc}, t_f^{inc})$ (Line \ref{algline:init_incumbent}) with the lowest-cost trajectory found so far. Whenever the low-level search finds a trajectory executing a tour $\Gamma$, it adds $\Gamma$ to the \textit{forbidden set} $\mathcal{H}_{forbid}$ (Line \ref{algline:init_H_forbid}), forbidding the GTSP-TW solver from producing $\Gamma$ again. On the other hand, if the low-level search fails to find a trajectory for a tour $\Gamma$, it adds the shortest prefix of $\Gamma$ that cannot be executed to $\mathcal{H}_{forbid}$.

FMC*-TSP ensures bounded-suboptimality by terminating only when the incumbent satisfies $t_f^{inc} \leq wLB$, where $LB$ is a lower bound on the optimal cost $t_f^*$. $LB$ is the minimum of two values, $LB_H$ and $LB_L$. $LB_H$ is a value maintained by the GTSP-TW solver, lower bounding the cost of any tour the solver can currently produce. $LB_L$ is a value updated by the low-level search. In particular, whenever the low-level search produces a trajectory for a tour $\Gamma$, it also produces a lower bound on the cost of executing $\Gamma$, and $LB_L$ is continually updated to be the lowest of these lower bounds.

Finally, for pruning purposes, FMC*-TSP maintains a dictionary called $\mathfrak{D}$. The keys for $\mathfrak{D}$ are tuples $(\mathcal{S}, u)$, where $\mathcal{S} \subseteq \mathcal{I}$, and $u$ is a target-window. The value for a key $(\mathcal{S}, u)$ is the cost of the best trajectory found so far intercepting the targets in $\mathcal{S}$, in any order, and terminating by intercepting $u$. We define the function $\Key(\Gamma)$, which converts a partial tour $\Gamma$ into a key $(\mathcal{S}, \Gamma[-1])$, where $S$ contains the targets visited by $\Gamma$. Whenever we generate a trajectory for a partial tour $\Gamma$, we constrain the trajectory to execute $\Gamma$ within time $\mathfrak{D}[\Key(\Gamma)]$ and prune $\Gamma$ if this is infeasible. If we attempt to access $\mathfrak{D}[(S, u)]$ for some key $(S, u)$ not contained in $\mathfrak{D}$, we assume $\mathfrak{D}[(S, u)]$ evaluates to $\overline{t}(u)$.

\vspace{-0.4cm}
\begin{algorithm}\label{alg:fmc_star_tsp}
\caption{FMC*-TSP}

\SetKwFunction{LowLevelSearch}{LowLevelSearch}
\SetKwFunction{UpdateUpperLimit}{UpdateUpperLimit}
\SetKwFunction{ConstructTargetWindowGraph}{ConstructTargetWindowGraph}
\SetKwFunction{ExcludePartialTour}{ExcludePartialTour}
\SetKwFunction{SolveGTSPTW}{SolveGTSP-TW}
\SetKwProg{Fn}{Function}{:}{}
\SetKwComment{Comment}{// }{}

$\mathcal{G}_{tw}$ = \ConstructTargetWindowGraph()\;
$(\Gamma^{inc}, \tau_a^{inc}, t_f^{inc}) = (NULL, NULL, \infty)$\;\label{algline:init_incumbent}
$\mathcal{H}_{forbid} = \{\}$\;\label{algline:init_H_forbid}
$LB_{L} = \infty$\;\label{algline:LB_L_init}
$\mathfrak{D}$ = dict()\;
\SolveGTSPTW($\mathcal{G}_{tw}$, \LowLevelSearch)\;
\lIf{\upshape $t_f = \infty$}{return INFEASIBLE}\label{algline:term_infeas}
return $(\Gamma^{inc}, \tau_a^{inc}, t_f^{inc})$\;
\end{algorithm}
\vspace{-0.65cm}

\subsection{Target-Window Graph}\label{subsec:twg}
We define a graph $\mathcal{G}_{tw} = (\mathcal{V}_{tw}, \mathcal{E}_{tw})$, called a \textit{target-window graph}. $\mathcal{V}_{tw}$ is the set of nodes, containing all target-windows. $\mathcal{E}_{tw}$ is the set of edges. To define the edges, we need the following quantities:

\begin{definition}
Given target-windows $u$ and $v$, the \textnormal{obstacle-unaware latest feasible departure time} $\mathbf{l}_{uv}$ is the maximum $t \in [\underline{t}(u), \overline{t}(u)]$ such that a speed-admissible trajectory exists intercepting $u$ at some $t_0 \in [\underline{t}(u), \overline{t}(u)]$, then $v$ at $t$, in the absence of obstacles. If no such trajectory exists, $\mathbf{l}_{uv} = -\infty$.
\end{definition}

\begin{definition}
Given target-windows $u$ and $v$ and departure time $t_0$ from $u$, the \textnormal{obstacle-unaware earliest feasible arrival time} $\mathbf{e}_{uv}(t_0)$ is the minimum $t \in [\underline{t}(v), \overline{t}(v)]$ such that a speed-admissible trajectory exists intercepting $u$ at time $t_0$ then $v$ at time $t$, in the absence of obstacles. If no such trajectory exists, $\mathbf{e}_{uv}(t_0) = \infty$. Additionally, when we write $\mathbf{e}_{uv}$ without any argument, we imply the argument $t_0 = \underline{t}(u)$.
\end{definition}

\begin{definition}
Given target-windows $u$ and $v$, the \textnormal{obstacle-unaware shortest feasible travel time} $\mathbf{s}_{uv}$ equals $\min_{t_0 \in [\underline{t}(u), \overline{t}(u)]} (\mathbf{e}_{uv}(t_0) - t_0)$. If $\mathbf{e}_{uv} = \infty$, $\mathbf{s}_{uv} = \infty$.
\end{definition}
$\mathbf{l}_{uv}$, $\mathbf{e}_{uv}$, and $\mathbf{s}_{uv}$ have closed-form expressions \cite{philip2023CStar}. For all pairs of target-windows $(u, v)$ with $\Tar(u) \neq \Tar(v)$, we first compute $\mathbf{l}_{uv}$; if $\mathbf{l}_{uv} \neq -\infty$, we compute $\mathbf{e}_{uv}$ and $\mathbf{s}_{uv}$, then store $\mathbf{l}_{uv}, \mathbf{e}_{uv}$ and $\mathbf{s}_{uv}$ in an edge $(u, v) \in \mathcal{E}_{tw}$. If $\mathbf{l}_{uv} = -\infty$, then $(u, v) \notin \mathcal{E}_{tw}$. For edge $e  = (u, v) \in \mathcal{E}_{tw}$, let $\mathbf{l}_e = \mathbf{l}_{uv}$, $\mathbf{e}_{e} = \mathbf{e}_{uv}$ and $\mathbf{s}_e = \mathbf{s}_{uv}$.

\subsection{GTSP-TW}\label{subsec:gtsptw}
We solve a GTSP-TW on the target-window graph for a tour, using the mixed integer linear program (MILP) formulation from \cite{yuan2020branch}, modified to incorporate $\mathbf{l}_{e}, \mathbf{e}_{e}$, and $\mathbf{s}_{e}$ values. The decision variables consist of a binary variable $x_{e} \in \{0, 1\}$ for each edge $e \in \mathcal{E}_{tw}$, indicating whether $e$ is traveled in the tour, as well as an arrival time $t^i$ for each $i \in \mathcal{I} \cup \{0\}$. For a target-window $v$, $\delta^-(v)$ is the set of edges entering $v$, and $\delta^+(v)$ is the set of edges leaving $v$. Let $\delta(i, j) = \bigcup\limits_{u \in \mathcal{T}_i} \bigcup\limits_{v \in \mathcal{T}_j}(\delta^+(u) \cap \delta^-(v))$. The MILP is given in Problem \ref{optprob:gtsp_tw}.
\begin{mini!}
{\substack{\{x_{e}\}_{e \in \mathcal{E}_{tw}},\\\{t^i\}_{i \in \mathcal{I} \cup \{0\}}}}{t^0\label{eqn:gtsp_tw_obj}}{\label{optprob:gtsp_tw}}{}
\addConstraint{\sum\limits_{v \in \mathcal{T}_i}\sum\limits_{e \in \delta^-(v)}x_e = 1 \; \forall i \in \mathcal{I} \cup \{0\}\label{eqn:gtsp_tw_visit_all_targ}}{}
\addConstraint{\sum\limits_{e \in \delta^-(v)}x_e = \sum\limits_{e \in \delta^+(v)}x_e \; \forall v \in \mathcal{V}_{tw}\label{eqn:gtsp_tw_flow_conservation}}{}
\addConstraint{\sum\limits_{v \in \mathcal{T}_i}\sum\limits_{e \in \delta^-(v)}\hspace{-0.25cm}\mathbf{e}_ex_e \leq t^{i} \leq \sum\limits_{u \in \mathcal{T}_i}\sum\limits_{e \in \delta^+(u)}\hspace{-0.25cm}\mathbf{l}_ex_e \; \forall i \in \mathcal{I}\label{eqn:gtsp_tw_time_bounds}}{}
\addConstraint{[i \neq 0]t^i - t^j + \hspace{-0.2cm}\sum\limits_{e \in \delta(i, j)}\mathbf{s}_ex_e \leq \sum\limits_{u \in \mathcal{T}_i}\sum\limits_{e \in \delta^+(u)}\hspace{-0.1cm}\mathbf{l}_ex_e\nonumber}{}
\addConstraint{-\sum\limits_{v \in \mathcal{T}_j}\sum\limits_{e \in \delta^-(v)}\mathbf{e}_ex_e - \sum\limits_{e \in \delta(i, j)}(\mathbf{l}_e - \mathbf{e}_e)x_e\nonumber}{}
\addConstraint{\phantom{\sum\limits_{v \in \mathcal{T}_j}\sum\limits_{e \in \delta^-(v)}\mathbf{e}_ex_e - } \forall i \in \mathcal{I} \cup \{0\}, j \in \mathcal{I}\label{eqn:gtsp_tw_travel_time}}{}
\addConstraint{t^i + \sum\limits_{e \in \delta^-(u_d)}\mathbf{s}_ex_e \leq t^0 \; \forall i \in \mathcal{I}\label{eqn:gtsp_tw_travel_time_depot}}{}
\addConstraint{\sum\limits_{e \in \text{edges}(\Gamma)}x_e \leq |\Gamma| - 2 \; \forall \Gamma \in \mathcal{H}_{forbid}\label{eqn:gtsp_tw_exclude}}{}
\end{mini!}
\eqref{eqn:gtsp_tw_obj} minimizes the arrival time at the depot. \eqref{eqn:gtsp_tw_visit_all_targ} requires the tour to visit all targets and the depot. \eqref{eqn:gtsp_tw_flow_conservation} requires that if the tour arrives at a target-window $v$, the tour also departs from $v$. \eqref{eqn:gtsp_tw_time_bounds} expresses that the arrival time at target $i$ is no earlier than $\mathbf{e}_e$ for the edge $e$ chosen to arrive at $i$, and that the arrival time is no later than $\mathbf{l}_e$ for the edge $e$ chosen to depart from $i$. \eqref{eqn:gtsp_tw_travel_time} ensures that if a tour contains $j$ immediately after $i$, the arrival time at $j$ is no earlier than the arrival time at $i$ plus $\mathbf{s}_e$ for the edge chosen to travel from $i$ to $j$. In \eqref{eqn:gtsp_tw_travel_time}, $[i \neq 0]$ equals 1 if $i \neq 0$ and equals 0 otherwise, enforcing departure from the depot at $t = 0$. \eqref{eqn:gtsp_tw_travel_time_depot} enforces a similar constraint to \eqref{eqn:gtsp_tw_travel_time} for traveling to the depot. \eqref{eqn:gtsp_tw_exclude} requires that any partial tour $\Gamma$ in $\mathcal{H}_{forbid}$ is not a prefix of a returned solution\footnote{For a partial tour $\Gamma = (u^1, u^2, \dots, u^{|\Gamma|})$, we define $\text{edges}(\Gamma) = \{(u^1, u^2), (u^2, u^3), \dots, (u^{|\Gamma| - 1}, u^{|\Gamma|})\}$}.

\subsection{Low-Level Search}\label{subsec:low_level_search}
When solving the MILP \ref{optprob:gtsp_tw}, a standard solver will produce several tours. For each tour $\Gamma$, the low-level search (LLS) attempts to generate a trajectory executing $\Gamma$, as well as a lower bound $\underline{g}(\Gamma)$ on the cost to execute $\Gamma$. LLS does so by computing a trajectory and lower bound for each prefix of $\Gamma$ that it has not seen before, in increasing order of prefix length, using FMC* (Section \ref{subsec:fmc_star}). If trajectory computation fails for a prefix $\Gamma[:i]$ (i.e. FMC* returns $\underline{g}(\Gamma[:i]) = \infty$), we add $\Gamma[:i]$ to $\mathcal{H}_{forbid}$ and return to the high-level search. If we successfully generate a trajectory for $\Gamma$, we update the incumbent with $\Gamma$, and we update $LB_L$ using $\underline{g}(\Gamma)$.

\vspace{-0.4cm}
\begin{algorithm}\label{alg:low_level_search}
\caption{\protect\LowLevelSearch{$\Gamma$}}

\SetKwFunction{SawBefore}{SawBefore}
\SetKwFunction{FMCStar}{FMC*}
\SetKwProg{Fn}{Function}{:}{}
\SetKwComment{Comment}{// }{}

\For{\upshape $i$ in ($2, 3, \dots, |\Gamma|$)}{
    \lIf{\SawBefore$(\Gamma[:i])$}{continue}
    
    $(\underline{g}(\Gamma[:i]), \tau_a, t_f)$ = \FMCStar($\Gamma[:i]$)\;
    $\mathcal{H}_{forbid} \leftarrow \mathcal{H}_{forbid} \cup \{\Gamma[:i]\}$\;
    \lIf{\upshape $\underline{g}(\Gamma[:i]) = \infty$}{
        return
    }
}
$(\Gamma^{inc}, \tau_a^{inc}, t_f^{inc}) \leftarrow (\Gamma, \tau_a, t_f)$\;
$LB_L \leftarrow \min(LB_L, \underline{g}(\Gamma))$\;
\end{algorithm}
\vspace{-0.6cm}

\subsection{FMC*}\label{subsec:fmc_star}
The low-level search invokes FMC* with a partial tour $\Omega$ (the same as $\Gamma[:i]$ in Alg. \ref{alg:low_level_search}) and seeks a trajectory executing $\Omega$, as well as a lower bound on its execution cost $\underline{g}(\Omega)$. Before invoking FMC* for the first time, we decompose the obstacle-free configuration space $\Qfree$ into convex regions $\{\mathcal{A}^1, \mathcal{A}^2, \dots, \mathcal{A}^{n_{reg}}\}$. Since the obstacle maps in our experiments are cubic grids, we choose to decompose $\Qfree$ into axis-aligned bounding boxes, but other decomposition methods, e.g. \cite{deits2015computing}, can be used as well. We then define a graph of convex sets (GCS), denoted as $G_{cs} = (\mathcal{V}_{cs}, \mathcal{E}_{cs})$, where $\mathcal{V}_{cs}$ is the set of nodes and $\mathcal{E}_{cs}$ is the set of edges. Each node in $\mathcal{V}_{cs}$ is a convex subset of $\mathcal{Q} \times \mathbb{R}$, i.e. a set of configurations and times. For each convex region $\mathcal{A}$ decomposing $\Qfree$, we have a \textit{region-node} $\mathcal{X}_\mathcal{A} = \mathcal{A} \times [-\infty, \infty]$ in $\mathcal{V}_{cs}$. For each target-window $u$, we have a \textit{window-node} $\mathcal{X}_u \in \mathcal{V}_{cs}$ containing the positions and times along $\Traj(u)$ within time interval $[\underline{t}(u), \bar{t}(u)]$: this set of positions and times is a line segment in space-time, and thereby a convex set \cite{philip2024mixedinteger}. We say an agent trajectory \textit{intercepts} a window-node if it intercepts the associated target-window. An edge connects $\mathcal{X}$ to $\mathcal{X}'$ if $\mathcal{X} \cap \mathcal{X}' \neq \emptyset$. We call a sequence of GCS nodes $P$ a \textit{path}. Similar to the notation in Section \ref{sec:problem_setup}, $|P|$ is the length of $P$, $P[j]$ is the $j$th element of $P$, and $P[-1]$ is the final element of $P$. We say path $P$ is a \textit{solution path} for $\Omega$ if $(\mathcal{X}_{\Omega[1]}, \mathcal{X}_{\Omega[2]}, \dots \mathcal{X}_{\Omega[-1]})$ is a subsequence of $P$. FMC* interleaves the construction of a solution path and the optimization of an associated trajectory.

FMC* finds a path and trajectory using focal search. The search maintains two priority queues, OPEN and FOCAL, each containing paths. Each path $P \in \text{OPEN}$ has an $f$-value $f(P)$, lower bounding the cost of a solution path beginning with $P$. We construct $f(P)$ as the sum of a cost-to-come $g(P)$ and a cost-to-go $h(P)$. OPEN prioritizes paths with smaller $f$-values. Let $f_{min}(\text{OPEN}) = \min_{v \in \text{OPEN}}f(v)$, and let $f_{min}(\text{OPEN}) = \infty$ if OPEN is empty. $f_{min}(\text{OPEN})$ is a lower bound on the cost of executing $\Omega$. When FMC* returns, along with returning a trajectory, it returns $f_{min}(\text{OPEN})$, which becomes $\underline{g}(\Gamma[:i])$ in Alg. \ref{alg:low_level_search}. In addition to its $f$-value, a path $P$ also stores an obstacle-free, speed-admissible agent trajectory $\LTraj(P)$ and a time $\LTime(P)$, such that $\LTraj(P)$ intercepts all window-nodes in $P$ in order, and $\LTime(P)$ is the interception time of the last window-node in $P$ ("L" stands for last).

The priority queue FOCAL stores all $P \in \text{OPEN}$ with $f(v) \leq wf_{min}(\text{OPEN})$. Each $P \in \text{FOCAL}$ also has a priority vector $\vec{f}_{pr}(P) = [\text{unv}(P), g + wh(P)]^T$, where $\text{unv}(P)$ is the number of target-windows in $\Omega$ without an associated GCS node in $P$, i.e. unvisited by $P$. $g + wh(P)$ is the priority function from weighted A* \cite{pohl1970heuristic}. FOCAL prioritizes nodes with lexicographically smaller priority vectors, i.e. it prefers paths with fewer unvisited target-windows, and for two paths with the same number of unvisited target-windows, FOCAL prioritizes them in the same way as weighted A*.

When invoking FMC* in Alg \ref{alg:low_level_search}, note that if $|\Omega| > 2$, we must have already computed a trajectory for $\Omega[:|\Omega| - 1]$. We can reuse the OPEN list from this previous search to speed up the current one (Line \ref{algline:reuse_open}), and this reuse is described in Section \ref{subsubsec:ll_search_reuse}. If $|\Omega| = 2$, we cannot reuse a previous OPEN list, so we initialize OPEN as empty, then add a path to OPEN containing only the depot (Lines \ref{algline:init_open_scratch_begin}-\ref{algline:init_open_scratch_end}).

Each search iteration pops a path $P$ from FOCAL, then obtains $\NextWindowIndex(\Omega, P)$, which is the smallest index $n$ such that $\mathcal{X}_{\Omega[n]} \notin P$. Next, for any $j \in \{n, n + 1, \dots |\Omega|\}$, we check if $\Key(\Omega[:j])$ is in $\mathfrak{D}$, and if so, we check if $\mathfrak{D}[\Key(\Omega[:j])]$ is equal to the start of the time window for $\Omega[j]$ (Line \ref{algline:ll_check_limit_is_start}). If so, there is no use finding a trajectory for $\Omega[:j]$, and we prune $P$\footnote{In practice, to account for numerical tolerances, we use the condition $\mathfrak{D}[\Key(\Omega[:j])] \leq \underline{t}(\Omega[j]) + \epsilon$, with $\epsilon = 10^{-8}$.}. If we have not pruned $P$, we obtain the \textit{successor nodes} of $P$. $\mathcal{X}' \in \mathcal{V}_{cs}$ is a successor node of $P$ if the following conditions hold:
\begin{enumerate}
    \item $(P[-1], \mathcal{X}') \in \mathcal{E}_{cs}$
    \item $\mathcal{X}'$ does not occur in $P$ after the final window-node in $P$, ensuring that descendants of $P$ will not visit a node more than once between visits to two window-nodes
    \item If $\mathcal{X}'$ is a window-node, $\mathcal{X}' = \mathcal{X}_{\Omega[n]}$
\end{enumerate}
From each successor node $\mathcal{X}'$ of $P$, we create a \textit{successor path} $P'$ by appending $\mathcal{X}'$ to $P$.

\begin{figure}
    \centering
    \vspace{0.2cm}
    \includegraphics[width=0.48\textwidth]{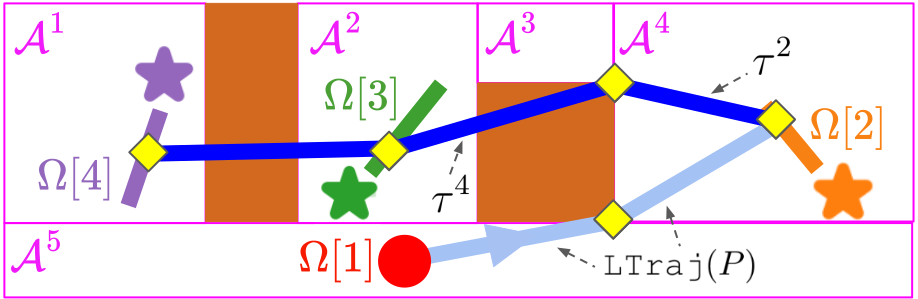}
    \vspace{-0.7cm}
    \caption{Computing $f(P')$ for $P' = (\mathcal{X}_{\Omega[1]}, \mathcal{X}_{\mathcal{A}^5}, \mathcal{X}_{A^4}, \mathcal{X}_{\Omega[2]}, \mathcal{X}_{\mathcal{A}^4}. \mathcal{X}_{\mathcal{A}^3})$. $P_{aux} = (\mathcal{X}_{\Omega[2]}, \mathcal{X}_{\mathcal{A}^4}, \mathcal{X}_{\mathcal{A}^3}, \mathcal{Q} \times \mathbb{R})$. Regions decomposing $\Qfree$ are shown as pink boxes. Trajectory segment endpoints are shown as yellow diamonds. $\tau^1$ is a zero-length segment between $\protect\LTraj(P)$ and $\tau^2$, and $\tau^3$ is a zero-length segment between $\tau^2$ and $\tau^4$. Concatenating $\protect\LTraj(P)$ with $\tau^1, \tau^2$, and $\tau^3$ gives an obstacle-free trajectory $\tau'$ with cost $g(P')$. $\tau^4$ and its extension ignore obstacles. Extension of $\tau^4$ terminates on purple target at time $f(P')$.}
    \label{fig:fmc_star_fig}
    \vspace{-0.6cm}
\end{figure}

We now describe how to compute $f(P')$ and $\vec{f}_{pr}(P')$. The procedure is illustrated in Fig. \ref{fig:fmc_star_fig}. We first obtain the index of the final window-node in $P$: call this index $j$. We then construct an auxiliary path, $P_{aux} = (P[j], P[j + 1], \dots, P[-1], \mathcal{Q} \times \mathbb{R})$. Next, we optimize a speed-admissible trajectory, parameterized as a sequence of contiguous linear trajectory segments $(\tau^1, \tau^2, \dots, \tau^{|P_{aux}|})$ and a sequence of segment end times $(t^1, t^2, \dots, t^{|P_{aux}|})$, such that $t^{|P_{aux}|}$ is minimized, segment $k$ lies entirely in set $P_{aux}[k]$, $\tau^1(\LTime(P)) = \LTraj(P)(\LTime(P))$, and $\tau^{|P_{aux}|}(t^{|P_{aux}|}) = \Traj(\Omega[n])(t^{|P_{aux}|})$. This optimization is the same as \cite{natarajan2024ixg}, eqn. 2, with the additional initial and terminal constraints. If the optimal final time $t^{|P_{aux}|}$ (denoted as $t$ on Line \ref{algline:trajopt}) is larger than $\mathfrak{D}[\Key(\Omega[:n])]$, we prune $P'$. Otherwise, we note that if we concatenate $\LTraj(P)$ with segments $k = 1$ to $k = |P_{aux}| - 1$, we obtain an obstacle-free trajectory (denoted as $\tau'$ on Line \ref{algline:trajopt}). The cost of $\tau'$ is $t^{|P_{aux}| - 1}$, and we let $g(P') = t^{|P_{aux}| - 1}$. If $P'[-1] = \mathcal{X}_{\Omega[n]}$, $\tau'$ is a new trajectory $\tau_{\Omega[:n]}$ executing $\Omega[:n]$, and we update $\mathfrak{D}[\Key(\Omega[:n])]$ (Line \ref{algline:ll_update_limit}) to prune future trajectories worse than $\tau_{\Omega[:n]}$. Next, we note that $\tau^{|P_{aux}|}$ travels in a straight line to position $\Traj(\Omega[n])(t^{|P_{aux}|})$, ignoring obstacles, terminating at time $t^{|P_{aux}|}$, denoted as $t$ on Line \ref{algline:trajopt}. Lines \ref{algline:efat_chain}-\ref{algline:efat_chain_done} extend $\tau^{|P_{aux}|}$ to intercept all remaining target-windows $\Omega[j]$ in $\Omega$ by successively computing the obstacle-unaware earliest feasible arrival time from one target-window to the next. If the arrival time at any $\Omega[j]$ is larger than $\mathfrak{D}[\Key(\Omega[:j])]$, we prune $P’$. Otherwise, $f(P')$ is the arrival time at $\Omega[-1]$, and $h(P') = f(P') - g(P')$, letting us compute $\vec{f}_{pr}$.

\vspace{-0.4cm}
\begin{algorithm}\label{alg:fmc_star}
\caption{\protect\FMCStar($\Omega$)}
\SetKwFunction{FMCStar}{FMC*}
\SetKwFunction{OptimizeTrajectory}{OptimizeTrajectory}
\SetKwFunction{SuccessorNodes}{SuccessorNodes}
\SetKwFunction{ReuseOpen}{ReuseOpen}
\SetKwFunction{UpdateFOCAL}{UpdateFOCAL}
\SetKwFunction{FoundEquivTraj}{FoundEquivTraj}
\SetKwFunction{Append}{Append}
\SetKwProg{Fn}{Function}{:}{}
\SetKwComment{Comment}{// }{}

\eIf{\upshape $|\Omega| = 2$}{\label{algline:init_open_scratch_begin}
    OPEN = []\;\label{algline:ll_init_open1}
    Push $P = (\mathcal{X}_{u_d})$ onto OPEN with $f(P) = 0$\;\label{algline:init_open_scratch_end}
}{
    OPEN = \ReuseOpen($\Omega[:|\Omega| - 1]$, $\Omega$)\;\label{algline:reuse_open}
}
$\tau_{\Omega}$ = NULL\;
\While{\upshape OPEN is not empty and ($\tau_{\Omega}$ is NULL or $\mathfrak{D}[\Key(\Omega)] > wf_{min}(\text{OPEN})$}{\label{algline:check_open_empty_ll}
    FOCAL = $\{P \in \text{OPEN} : f(P) \leq f_{min}(\text{OPEN})\}$\;
    $P$ = FOCAL.pop()\;\label{algline:pop_ll}
    OPEN.remove($P$)\;

    $n$ = \NextWindowIndex($\Omega, P$)\;\label{algline:next_window_index}
    \lIf{\upshape $\exists j \in \{n, n + 1, \dots |\Omega|\}$ s.t. $\Key(\Omega[:j])$ in $\mathfrak{D}$ and \upshape $\mathfrak{D}[\Key(\Omega[:j])] = \underline{t}(\Gamma[j])$}{
        \hspace{-0.15cm}continue\hspace{-0.1cm}\label{algline:ll_check_limit_is_start}
    }

    \For{\upshape $\mathcal{X}'$ in \SuccessorNodes($P$)}{\label{algline:get_succ_sets}
        $P' = \Append(P, \mathcal{X}')$\;
        
        $t, g(P'), \tau'$ = \OptimizeTrajectory($P'$)\;\label{algline:trajopt}

        \lIf{\upshape $t > \mathfrak{D}[\Key(\Omega[:n])]$}{
            continue\label{algline:ll_infeas}
        }

        \eIf{\upshape $\mathcal{X}' = \mathcal{X}_{\Omega[n]}$}{
            $\tau_{\Omega[:n]} = \tau'$, $\mathfrak{D}[\Key(\Omega[:n])] = g(P')$\;\label{algline:ll_update_limit}
            $\LTraj(P') = \tau'$, $\LTime(P') = g(P')$\;
        }{
            $\LTraj(P') = \LTraj(P)$\;
            $\LTime(P') = \LTime(P)$\;
        }

        \For{\upshape $j$ in ($n + 1, n + 2, \dots, |\Omega|$)}{\label{algline:efat_chain}
            $t \leftarrow \mathbf{e}_{\Omega[j - 1]\Omega[j]}(t)$\;
            \lIf{$t > \mathfrak{D}[\Key(\Omega[:j])])$}{
                break
            }
        }
        \lIf{\upshape $t > \mathfrak{D}[\Key(\Omega[:j])])$}{
            continue\label{algline:efat_chain_done}
        }
        Push $P'$ onto OPEN with $f(P') = t$\;
        $h(P') = f(P') - g(P')$\;\label{algline:ll_h_val}
        $\vec{f}_{pr}(P') = [\text{unv}(P'), g(P') + wh(P')]^T$\;\label{algline:f_pr}
    }
}
return $f_{min}(\text{OPEN})$, $\tau_{\Omega}, \mathfrak{D}[\Key(\Omega)]$\;
\end{algorithm}
\vspace{-0.5cm}

\subsubsection{Reusing OPEN Between FMC* Calls}\label{subsubsec:ll_search_reuse}
When invoking FMC* with $|\Omega| > 2$, we set OPEN equal to the OPEN list at the end of the FMC* search for $\Omega[:|\Omega| - 1]$, with updated $f$-values and $\vec{f}_{pr}$ vectors. For each $P \in \text{OPEN}$, before updating $f(P)$ and $\vec{f}_{pr}(P)$, we execute Lines \ref{algline:next_window_index}-\ref{algline:ll_check_limit_is_start} and prune $P$ if the condition on Line \ref{algline:ll_check_limit_is_start} holds. If we did not prune $P$, we perform the update $f(P) \leftarrow \mathbf{e}_{\Omega[-2]\Omega[-1]}(f(P))$. We then update $\vec{f}_{pr}(P)$ by executing lines \ref{algline:ll_h_val}-\ref{algline:f_pr} with $P' = P$.

\section{THEORETICAL ANALYSIS}
In this section, we sketch proofs for the bounded-suboptimality and completeness of FMC*-TSP.

\begin{theorem}\label{thm:bounded_subopt}
For a feasible problem instance, Alg. \ref{alg:fmc_star_tsp} returns a solution with cost no more than $w$ times the optimal cost.
\end{theorem}
Let $\Gamma^*$ be a tour whose minimum execution cost is equal to the optimal cost of a problem instance, $t_f^*$. At any time, either $\Gamma^*$ corresponds to a feasible solution to Problem \ref{optprob:gtsp_tw}, meaning $LB_H$ lower bounds the execution cost of $\Gamma^*$, or the low-level search found a trajectory for $\Gamma^*$, meaning that $LB_L$ lower bounds the execution cost of $\Gamma^*$. This means $LB = \min(LB_H, LB_L) \leq t^*_f$. By terminating only when the incumbent satisfies $t_f^{inc} \leq wLB$, we ensure that $t_f^{inc} \leq wt_f^*$.

\begin{theorem}\label{thm:completeness_infeas}
For an infeasible problem instance, Alg. \ref{alg:fmc_star_tsp} will report that the instance is infeasible in finite time.
\end{theorem}
Finite-time termination follows from the finite number of tours Problem \ref{optprob:gtsp_tw} can generate and the finite number of paths that Alg. \ref{alg:fmc_star} can explore. Since no feasible solution exists, the incumbent cost $t_f^{inc}$ remains at its default value $\infty$, and Alg. \ref{alg:fmc_star_tsp} must return infeasible on Line \ref{algline:term_infeas}.

\section{NUMERICAL RESULTS}

\begin{figure*}[htb]
    \centering
    \vspace{0.2cm}
    \includegraphics[width=0.99\textwidth]{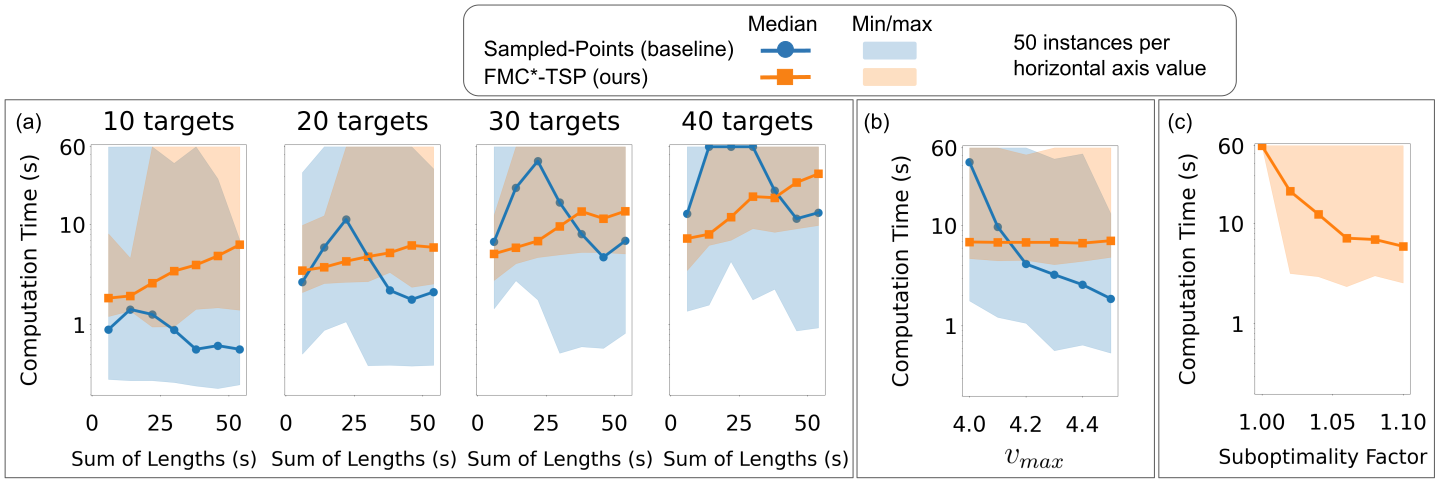}
    \vspace{-0.4cm}
    \caption{All vertical axes are on a log-scale. (a) Varying the number of targets and sum of time window lengths per target. (b) Varying the agent's speed limit. (c) Varying FMC*-TSP's suboptimality factor.}
    \label{fig:vary_tw_len_and_subopt}
    \vspace{-0.5cm}
\end{figure*}
We ran experiments on an Intel i9-9820X 3.3GHz CPU with 128 GB RAM. Experiment 1 (Section \ref{subsec:experiment1}) varied the number of targets and the sum of their time window lengths, Experiment 2 (Section \ref{subsec:experiment2}) varied the agent's speed limit, and Experiment 3 (Section \ref{subsec:experiment3}) varied FMC*-TSP's suboptimality factor. All problem instances had 2 time windows per target. We generated 1400 total instances by extending the instance generation method from \cite{bhat2024AComplete} to 3D. We compared FMC*-TSP to a baseline based on related work \cite{Li2019RendezvousPlanning, philip2023CStar, stieber2022DealingWithTime}, which samples trajectories of targets into points and solves a generalized traveling salesman problem (GTSP) to find a sequence of points to visit. The baseline solves its GTSP via the integer program \cite{laporte1987generalized} without subtour elimination constraints. Subtours are only possible if two samples are identical, which we ensure never occurs. Costs between sample points are initially an obstacle-unaware cost, and whenever the GTSP solver generates a sequence of points, we evaluate the obstacle-aware costs between consecutive pairs of points in parallel using a variant of IxG* \cite{natarajan2024ixg}. We limited each method's computation time to 60 s.

When we compared with the baseline (Experiments 1 and 2), we set a suboptimality factor $w = 1.1$ for FMC*-TSP. The baseline is not bounded-suboptimal for the MT-TSP-O, but it is bounded-suboptimal for the sampled approximation of the MT-TSP-O that it solves, and we set a suboptimality factor of 1.1 to make a roughly fair comparison with FMC*-TSP. The baseline is also not complete, but if it uses more sample points, it is more likely to find a solution. Therefore, we initialized the baseline with 5 points per target, and whenever the baseline failed to find a solution, we added 5 more points per target and tried again.

\subsection{Experiment 1: Varying Sum of Time Window Lengths}\label{subsec:experiment1}

We varied the number of targets from 10 to 40 and the sum of time window lengths per target from 6 s to 54 s. The results are in Fig. \ref{fig:vary_tw_len_and_subopt} (a). As we saw in \cite{bhat2024AComplete}, there is a range for the sum of window lengths where the baseline takes more median computation time than FMC*-TSP to find a solution, and this range widens as we increase the number of targets. For these ranges, the baseline needs an excessive number of sample points to find a feasible solution, since the fraction of a target's time windows that is part of a feasible solution is small. FMC*-TSP's computation time increases as we increase the sum of lengths, similar to previous results on TSP variants with time windows \cite{Dumas1995OptimalAlgorithm, philip2024mixedinteger}. Runtime increased in both major components of FMC*-TSP: the GTSP-TW mixed integer program and the low-level FMC* search. GTSP-TW runtime increases with time window length because when we have larger time windows, it is feasible more often to travel from one target-window to another, leading to more edges in the target-window graph and thereby more binary variables in Problem \ref{optprob:gtsp_tw}. FMC* computation time increases because FMC*-TSP calls FMC* more times: a larger number of edges in the target-window graph causes the GTSP-TW to produce more tours and seek trajectories for them from FMC*. When we increased the sum of lengths from 6 to 54, the median ratio of GTSP-TW time to FMC* time went from 0.062 to 0.18, i.e. while GTSP-TW time increased, FMC* remained the bottleneck.

In Table \ref{table:cost_comparison}, we show statistics for the percent difference in cost between FMC*-TSP and the baseline in cases where the both methods found a solution. Since we set FMC*-TSP's suboptimality factor to 1.1, its cost is never more than 10\% larger than the baseline's, and at best, its cost is 23\% lower.
\renewcommand{\arraystretch}{1.8}
\begin{table}
\caption{Comparison of the solution cost $t_{\text{FMC*-TSP}}$ from FMC*-TSP against the cost $t_\text{SP}$ from the sampled-points method.}\label{table:cost_comparison}
\vspace{-0.3cm}
\centering
\begin{tabular}{|@{\quad}l@{\quad}|@{\quad}l@{\quad}|@{\quad}l@{\quad}|@{\quad}l@{\quad}|}
\hline
 & Median & Min & Max \\
\hline
$\frac{t_\text{FMC*-TSP} - t_\text{SP}}{t_\text{SP}}*100\%$ & \textcolor{DarkGreen}{\textbf{-0.077\%}} & \textcolor{DarkGreen}{\textbf{-23\%}} & \textcolor{orange}{\textbf{9.9\%}}\\
\hline
\end{tabular}
\vspace{-0.5cm}
\end{table}

\vspace{-0.5cm}

\subsection{Experiment 2: Varying Agent's Speed Limit}\label{subsec:experiment2}
We varied the agent's speed limit $v_{max}$ from 4.0 to 4.5, fixing the number of targets to 30 and the sum of lengths per target to 22 s. The results are in Fig. \ref{fig:vary_tw_len_and_subopt} (b). As we increase $v_{max}$, the baseline's computation time decreases. This is because if the agent can move faster, the subintervals of targets' time windows that are part of a feasible solution grow larger, and it is more likely that a sample point lies within one of these subintervals.

\subsection{Experiment 3: Varying Suboptimality Factor}\label{subsec:experiment3}
We varied the suboptimality factor of FMC*-TSP from $w = 1.0$ to $w = 1.1$, fixing the number of targets to 20 and the sum of lengths to $54$ s. The results are shown in Fig. \ref{fig:vary_tw_len_and_subopt} (c). As expected, as the allowable suboptimality increases, FMC*-TSP computes solutions more quickly.

\section{CONCLUSION}
In this paper, we developed FMC*-TSP, a complete and bounded-suboptimal algorithm for the MT-TSP-O in 3D that leverages graphs of convex sets. We demonstrated a range of time window lengths for the targets where FMC*-TSP finds solutions more quickly than a baseline that samples trajectories of targets into points. A natural extension of this work would be to incorporate multiple agents.





\section*{ACKNOWLEDGMENT}
This material is partially based on work supported by the National Science Foundation (NSF) under Grant No. 2120219 and 2120529. Any opinions, findings, conclusions, or recommendations expressed in this material are those of the author(s) and do not necessarily reflect views of the NSF.

\FloatBarrier
\bibliographystyle{IEEEtran}
%
\bibliography{refs}

\clearpage

\end{document}